\documentclass[10pt,twocolumn,letterpaper]{article}

\usepackage{cvpr}              

\usepackage[dvipsnames]{xcolor}

\definecolor{cvprblue}{rgb}{0.21,0.49,0.74}
\usepackage[pagebackref,breaklinks,colorlinks,citecolor=cvprblue]{hyperref}

\usepackage[accsupp]{axessibility} 

\usepackage{hhline}
\usepackage{bbm}
\usepackage{amsmath}
\usepackage[utf8]{inputenc} %
\usepackage[T1]{fontenc}    %
\usepackage{hyperref}       %
\usepackage{url}            %
\usepackage{booktabs}       %
\usepackage{amsfonts}       %
\usepackage{nicefrac}       %
\usepackage{microtype}      %
\usepackage{xcolor}         %

\usepackage{graphicx}
\usepackage{multirow}
\usepackage{subcaption}
\usepackage{colortbl}
\usepackage{makecell, verbatim, sidecap}
\DeclareMathOperator*{\argmin}{argmin} %
\newcommand{\ours}{{\cellcolor[gray]{.9}}}

\def\paperID{9398} 
\def\confName{CVPR}
\def\confYear{2024}

\title{Rethinking Boundary Discontinuity Problem for Oriented Object Detection}

\author{
Hang Xu$^{1,2}$\thanks{Equal contribution}, Xinyuan Liu$^{2,3*}$, Haonan Xu$^{2,3}$ ,
Yike Ma$^{2}$, Zunjie Zhu$^{1,4}$ , Chenggang Yan$^{1}$, Feng Dai$^{2}$\thanks{Corresponding author} \vspace{2pt}\\
$^1$Hangzhou Dianzi University, Hangzhou, China \\
$^2$Institute of Computing Technology, Chinese Academy of Sciences, Beijing, China \\
$^3$University of Chinese Academy of Sciences, Beijing, China\\
$^4$Lishui Institute of Hangzhou Dianzi University, Lishui, China\\
{\tt\small
\{hxu, zunjiezhu, cgyan\}@hdu.edu.cn
\{liuxinyuan21s, xuhaonan23s, ykma, fdai\}@ict.ac.cn
}
}

\begin{document}


\maketitle
\begin{abstract}
Oriented object detection has been developed rapidly in the past few years, where rotation equivariance is crucial for detectors to predict rotated boxes. It is expected that the prediction can maintain the corresponding rotation when objects rotate, but severe mutation in angular prediction is sometimes observed when objects rotate near the boundary angle, which is well-known boundary discontinuity problem. 
The problem has been long believed to be caused by the sharp loss increase at the angular boundary, and widely used joint-optim IoU-like methods deal with this problem by loss-smoothing. However, we experimentally find that even state-of-the-art IoU-like methods actually fail to solve the problem. 
On further analysis, we find that the key to solution lies in encoding mode of the smoothing function rather than in joint or independent optimization. In existing IoU-like methods, the model essentially attempts to fit the angular relationship between box and object, where the break point at angular boundary makes the predictions highly unstable.
To deal with this issue, we propose a dual-optimization paradigm for angles. We decouple reversibility and joint-optim from single smoothing function into two distinct entities, which for the first time achieves the objectives of both correcting angular boundary and blending angle with other parameters.
Extensive experiments on multiple datasets show that boundary discontinuity problem is well-addressed. Moreover, typical IoU-like methods are improved to the same level without obvious performance gap. The code is available at \url{https://github.com/hangxu-cv/cvpr24acm}.
\end{abstract}  
\section{Introduction}
\label{sec:intro}

As an expansion of horizontal object detection \cite{ren2015faster,lin2017feature,lin2017focal}, oriented object detection has a wider applications in many scenes, such as aerial images \cite{ding2018learning,xu2020gliding}, panoramic images \cite{xu2022pandora, liu2023sph2pob,xu2023gldl}, scene text \cite{jiang2017r2cnn}, 3D objects \cite{zheng2020rotation}, etc, since it can achieve a good balance between fine localization and low labeling cost.
In oriented object detection, a detector needs to predict the minimal rotated  bounding boxes for objects, so it has a high requirement for rotation equivariance. However, researchers have observed mutation in angular prediction when objects rotate near the boundary angle, which is commonly known as boundary discontinuity problem \cite{yang2020arbitrary,yang2021rethinking}.

\begin{figure}[!tb] 
\centering
\includegraphics[width=\linewidth]{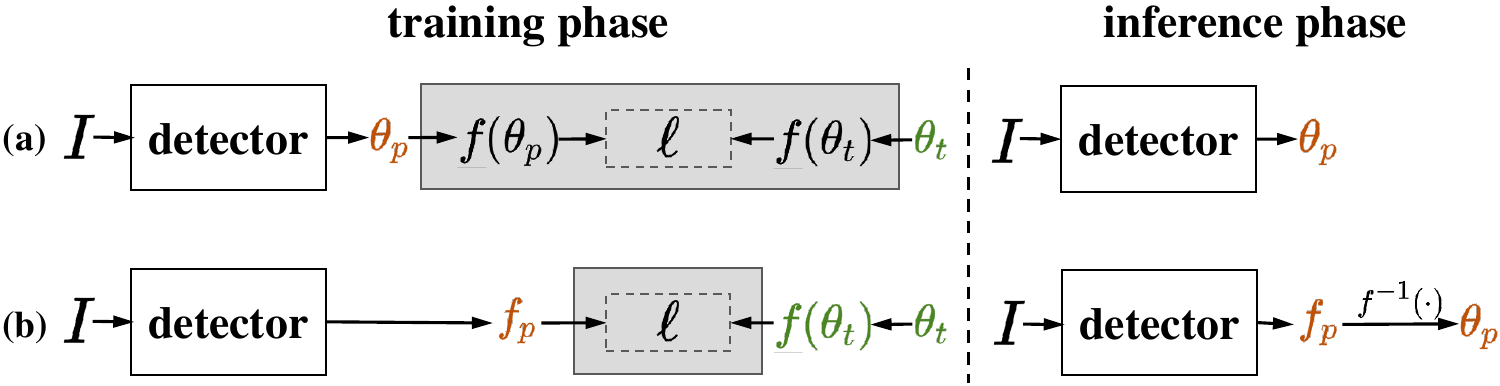}
\caption{Two optimization paradigms for angle in oriented object detection: (a) in joint-optim methods \cite{yang2021rethinking,yang2021learning,yang2023kfiou,zhou2019iou}, smoothing function is \textit{explicitly} applied for detector's output $\theta_p$ during loss calculation; (b) while in independent-optim methods \cite{yang2020arbitrary,yu2023psc}, smoothing function is \textit{implicitly} embedded in the model, and $\theta_p$ is decoded from detector's output $f_p$. According to our analysis, only the latter can really solve boundary discontinuity problem.}
\label{fig:key}
\end{figure}

\begin{figure*}
   \centering
   \includegraphics[width=0.88\linewidth]{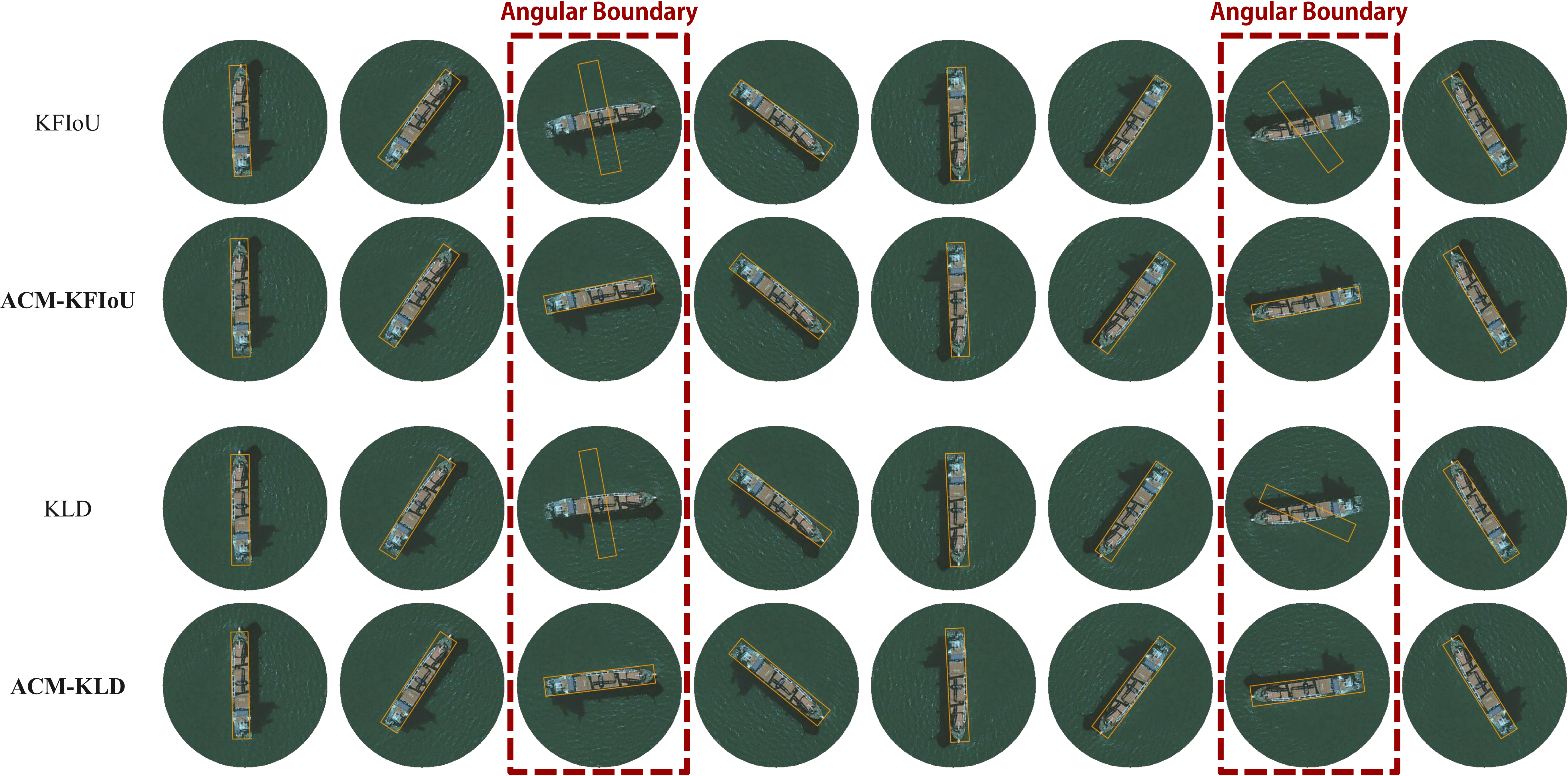}
   \caption{When objects rotate near the boundary angle,  state-of-the-art IoU-like methods (e.g., KFIoU \cite{yang2023kfiou}, KLD \cite{yang2021learning}) actually suffer from severe mutation in angular prediction. With the correction for angle by our ACM, the prediction achieves rotation equivariance.}
   \label{fig:error_case}
\end{figure*}

In previous works, the boundary discontinuity problem has been long believed to be caused by the sharp loss increase at the angular boundary during training. To address this problem, researchers designed a series of smooth loss functions to prevent the sharp loss increase, and these methods can be divided into two categories, i.e., independent-optim loss \cite{yang2020arbitrary,yang2021dense,yu2023psc} and joint-optim loss (dominated by IoU-like loss) \cite{yang2021learning,yang2021rethinking,zhou2019iou,yang2023kfiou}. Due to the negative impact of the low consistency between loss and IoU-metric, the detectors trained through the independent-optim loss are usually worse than IoU-like loss. It has long been a consensus in object detection \cite{yu2016unitbox,rezatofighi2019generalized,zheng2020distance}, so increasing IoU-like loss methods become mainstream choices for oriented object detectors.

However, we experimentally find that even state-of-the-art IoU-like methods do not actually solve the boundary discontinuity problem. Specifically, we select an image containing only a single object, and rotate it 360$^\circ$ at 1$^\circ$ intervals to obtain a series of images. These images are sequentially fed into a well-trained detector(with state-of-the-art IoU-like methods) for inference. As is shown in \cref{fig:error_case}, visualized results show that the predicted boxes can tightly enclose object in most cases, but collapse with a seriously deviated angle in some cases near the angular boundary.

Through theoretical analysis, we find that the key to addressing the problem lies in the encoding mode of the smoothing function rather than in joint or independent optimization. 
Although both optimization paradigms insisit loss-smoothing, the joint-optim methods have a subtle technical detail differing with independent-optim methods. As is shown as \cref{fig:key}, in joint-optim methods \cite{yang2021rethinking,yang2021learning,yang2023kfiou,zhou2019iou}, smoothing function is \textit{explicitly} applied for detector's output $\theta_p$ during loss calculation; while in independent-optim method \cite{yang2020arbitrary,yu2023psc}, smoothing function is \textit{implicitly} embedded in the model, and $\theta_p$ is decoded from detector's output $f_p$. For example, in typical joint-optim method KLD \cite{yang2021learning}, Gaussian distribution is transformed from predicted angle and other parameters, not directly output from the model.
It is this detail that makes those IoU-like methods not really solve boundary discontinuity problem as they expect, even though they indeed improve the overall detection performance with the benefit of joint optimization.
Specifically, the model still attempts to fit the angular relationship between box and object. The relationship is actually a piecewise function with a break point at the angular boundary as \cref{fig:waves}, which is difficult to fit for intrinsically continuous neural networks \cite{cybenko1989approximation,zhou2020universality,lu2017expressive}. It makes angles highly unstable near breakpoints, and results in the boundary discontinuity problem.
Such being the case, an intuitive idea occurs that lets the model output a Gaussian distribution. However, it is challenging to recover the original rotation angles of bounding boxes from Gaussian distributions. If we want to have one's cake and eat it too, we must find a coding function that simultaneously satisfies the smooth, joint, and reversible characteristics.

To deal with this issue, we propose a dual-optimization paradigm for angles as \cref{fig:overview}. We decouple reversibility and joint-optim from single smoothing function into two distinct entities $f$ and $g$. The former corrects angular boundary, while the latter blends angle with other parameters. In this paradigm, the model outputs angular encoding $f_p$, subject to explicit supervision. On this basis, another joint-optim $g$ is applied into decoded angle $\theta_{p}=f_p^{-1}$. Obviously, the role of $g$ can be played by existing joint-optim methods. However, given that $f^{-1}$ is involved in loss calculation, it is necessary to ensure that $f^{-1}$ is differentiable, which is not satisfied for lots of existing encoding. Inspired by the continuous encoding of PSC \cite{yu2023psc}, we propose a coding function based on the complex-exponential function, achieving the goal of differentiability of the inverse function. 
Finally, boundary discontinuity problem is well-addressed as \cref{fig:error_case}. \textbf{Overall, our contribution can be summarized as following:}

\begin{itemize}
    \item 
    We extract and induce the optimization logic of existing methods from mathematical perspective, for the first time
    clarifying the long-standing misunderstanding that IoU-like methods can solve boundary problem.
    
    \item 
    We propose a novel dual-optimization paradigm for angles, which for the first time achieves the objectives of both correcting angular boundary and blending parameters, 
    achieving rotational equivariance for detection.
    
    \item 
    Extensive experiments on multiple datasets show that boundary discontinuity problem is well-addressed. Moreover, typical IoU-like methods are improved to the same level without obvious performance gap.

\end{itemize}


\section{Related Works}
\subsection{Rotated Object Detection}
In oriented detection, the minimal enclosing rotated bounding box $(x, y, w, h, \theta)$ is adopted widely to represent an oriented object, where $(x, y)$ is center $position$, $(w, h)$ is $scale$ (i.e., width \& height) and $\theta$ is rotated $angle$ of box. There are many algorithms inherited from classic horizontal detection \cite{girshick2015fast,ren2015faster,lin2017feature,lin2017focal} to predict the rotated boxes, where ROI-Transformer \cite{ding2018learning}, SCRDet \cite{yang2019scrdet}, ReDet \cite{han2021redet} are two-stage mainstreamed methods, while DRN \cite{pan2020dynamic}, R$^3$Det \cite{yang2021r3det}, S$^2$A-Net \cite{han2021align} are single-stage methods. However, these detectors suffer from boundary discontinuity problems in varying degrees, as the issue itself is unrelated to the detectors.

\subsection{Boundary Discontinuity Problem}
\textbf{The boundary discontinuity problem has been a persistent challenge, requiring a comprehensive understanding of the antecedents and consequences of each milestone to grasp the essence of this paper.}
In horizontal detection, bbox-regression loss typically employs joint-optim IoU-Loss, which has reached a consensus without controversy.
Due to the complexity and non-differentiability of IoU calculation for rotated box, it was initially considered that IoU-Loss can not be available for oriented detection. Therefore, early methods in oriented detection usually used L1-Loss for each parameters $(x, y, w, h, \theta)$.

\textbf{CSL \cite{yang2020arbitrary} pointed out that using L1-Loss would lead to sharp increases in angle-regression loss at angle boundaries, termed "boundary discontinuity problem".} By using angle classification instead of angle regression, CSL avoids the intractable problem. Subsequently, a series of methods (e.g., DCL \cite{yang2021dense}\ /\ GF-CSL \cite{wang2022gaussian}\ / \ MGAR \cite{wang2022multigrained}) based on angle classification have sprung up.

GWD \cite{yang2021rethinking} argued that while CSL solved the "boundary discontinuity problem" caused by sharp loss increases, independently optimizing parameters was unreasonable. This is because IoU-Loss was already established as the best choice in horizontal detection. However, since rotated IoU is non-differentiable, GWD proposed a Gaussian-based joint-optim loss to approximately replace it. Hence, GWD claimed that it can address the "boundary discontinuity problem" and achieve joint optimization.
KLD \cite{yang2021learning} and KFIoU \cite{yang2023kfiou} inherit the advantages of GWD's Gaussian encoding, and improve it from distribution measurement. Due to the remarkable effect of these methods, more and more Gaussian methods have emerged, which indicates that joint-optim methods have become mainstream. \textbf{Notably, the perception of the "boundary discontinuity problem" remained limited to sharp loss increases up to this point.}

Recently, PSC \cite{yu2023psc} borrows phase-shift-coding from the field of communications to improve the performance of angle prediction. It uses continuous coding to avoid quantization errors in classification methods, but it still belongs to independent optimization. Notably, \textbf{PSC focuses on coding design without new insight about boundary discontinuity problem} (e.g., it explicitly mentioned that GWD/KLD solved the boundary problem).


\section{Preliminary}

\begin{figure}[!tb]
\centering
\begin{minipage}[b]{0.17\linewidth}
  \centering
  \begin{subfigure}[b]{\linewidth}
    \includegraphics[width=\linewidth]{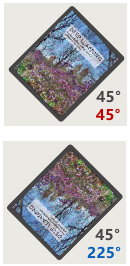}
    \caption{}
    \label{fig:object_box}
  \end{subfigure}
\end{minipage}
\hfill
\begin{minipage}[b]{0.78\linewidth}
  \centering
  \begin{subfigure}[b]{\linewidth}
    \includegraphics[width=\linewidth]{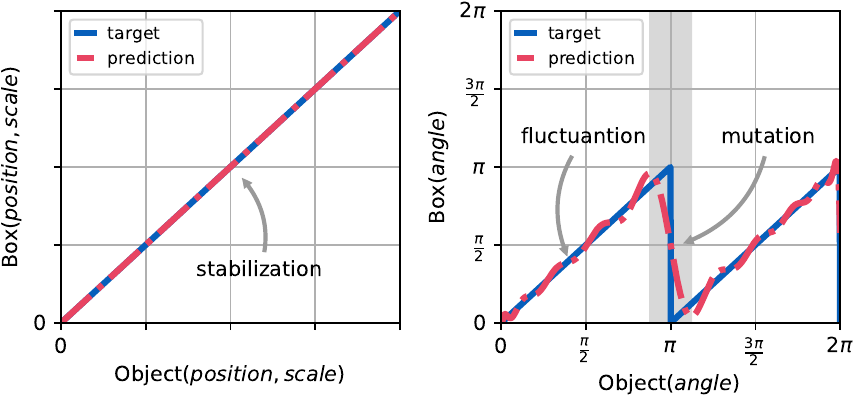}
    \caption{}
    \label{fig:waves}
  \end{subfigure}
\end{minipage}
\caption{Box $\neq$ Object: (a) objects rotated with 45$^\circ$
and 225$^\circ$[colorful mark] share the same box rotated with 45$^\circ$
[black mark], which causes (b) the relationship [blue line] between $angle$ of box and object to become a piecewise function with a breakpoint [gray region], differing from the $(position, scale)$. Not only is the prediction [red line] of the breakpoint region mutational, but the prediction of other regions also becomes fluctuant.}
\end{figure}

\subsection{The Root of All Evil is "Box $\neq$ Object"}
For an oriented object detector, it accepts image of object as input, and outputs bounding box with $position$, $scale$ and $angle$ parameters. However, we reveal that box and object are essentially different concepts, which will produce breakpoints in the angular ground-truth. The discontinuous ground-truth cannot be fitted exactly by continuous output of the detector especially at the breakpoints, so angular prediction near the breakpoints becomes very unstable. 

In the interest of brevity, we denote the object instance and bounding box as $O(x_{obj}, y_{obj}, w_{obj}, h_{obj}, \theta_{obj})$ and $B(x_{box}, y_{box}, w_{box}, h_{box}, \theta_{box})$, respectively. The difference between $O$ and $B$ lies in $\theta$ rather than $(x, y)$ and $(w, h)$, where the range of $\theta_{obj}$ is $[0, 2\pi)$ while the range of $\theta_{box}$ is $[0, \pi)$. This is because the object holds content which needs to rotate at least one full circle to be completely overlapped, while the box is a kind of geometry without any content which just needs to rotate half of circle to be completely overlapped. For example in \cref{fig:object_box}, objects rotated with $45^{\circ}$ and $225^{\circ}$ can be distinguished by content, while the corresponding bounding boxes cannot as well.

In this setting, the bounding box is a truly symmetric rectangle, whose rotations $\theta$ and $\theta \pm \pi$ are indistinguishable.
As a result, the relationship between $\theta_{box}$ and $\theta_{obj}$ exhibits a piecewise function with a break point, rather than a linear relationship between $(x_{box}, y_{box}, w_{box}, h_{box})$ and $(x_{obj}, y_{obj}, w_{obj}, h_{obj})$, as is shown in \cref{fig:waves} and \cref{eq:sawtooth}.
\begin{equation}
\begin{cases}
\begin{aligned}
(x_{box}, y_{box}) &= (x_{obj}, y_{obj}) \\
(w_{box}, h_{box}) &= (w_{obj}, h_{obj}) \\
\theta_{box} &= \theta_{obj} \bmod \pi
\end{aligned}
\end{cases}
\label{eq:sawtooth}
\end{equation}

The detector takes the object image as input and the box as supervision, which means that the detector is actually enforced to fit \cref{eq:sawtooth}(blue solid lines in \cref{fig:waves}). Obviously, $\theta_{box}$ has a step-point at $\theta_{obj} = \pi$, which makes it difficult for the detector $F$, a continuous function essentially, to fit it accurately. Irrespective of the quality of fit achieved by detector $F$, there always exists a small interval $(\pi-\epsilon_1, \pi+\epsilon_2)$ near the breakpoint (gray region in \cref{fig:waves}), where predicted angle (red dash line) drops rapidly from $\pi$ to $0$, and angular prediction becomes highly unstable, resulting in a severe degradation of the AP/IoU of boxes. In addition, angular prediction tends to fluctuate even outside the interval.

\subsection{The Devil is in Encoding Mode}
For the problem of angle discontinuity at the boundary, the core of the mainstream solutions is to smooth loss value at the angular boundary, and these studies are usually categorized by independent or joint optimization. 
However, our experiments as \cref{fig:error_case} shows that even joint-optim methods do not actually solve the boundary discontinuity problem. 

To understand the reason behind this finding, we make a reformulation of the existing works. For convenience, let the ground-truth and prediction of $\theta_{box}$ be denoted as $\theta_{t}$ and $\theta_{p}$, respectively. The way to optimize angle in joint-optim methods can be reformulated as follows (also as \cref{fig:key}):
\begin{equation}
\label{eq:loss-smoothing}
\theta 
= \mathop{\arg\min}\limits_{\theta_p} \ell \bigl (f(\theta_p); f(\theta_t) \bigr )
\end{equation}
where model fits discontinuous $\theta_p$, $f$ and $\ell$ are the encoding function for angle and measuring function for encoded value, respectively. 
For example, \textbf{1)} in the case of KLD \cite{yang2021learning}, $f = gaussian_{x, y, w, h}(\theta), \ell = \ell_{kld}$. $f$ encodes the angle and other parameters as a smooth Gaussian distribution, and $\ell$ just measures the distance of Gaussian distribution between prediction and ground-truth; \textbf{2)} in the case of SkewIoU \cite{zhou2019iou}, $f$ and $\ell$ are implicit functions derived from $SkewIoU(\theta_{p}^{xywh}, \theta_{t}^{xywh})$. Although we cannot get explicit expression of $f$ and $l$, their role must be similar to $gaussian_{x, y, w, h}(\theta)$ and $\ell_{kld}$.

As a contrast, the way to optimize angle in independent-optim methods can be formulated as follows (also as \cref{fig:key}):
\begin{align}
\label{eq:angle-smoothing}
\theta
&= f^{-1} \Bigl (\mathop{\arg\min}\limits_{f_p} \ell \bigl (f_p; f(\theta_t) \bigr ) \Bigr )
\end{align}
where model fits continous $f_p$, $f^{-1}$ is the inverse function of $f$, and we can get angle by $\theta_p = f^{-1}_{p}$. For example, \textbf{1)} in the case of CSL \cite{yang2020arbitrary}, $f=onehot(\theta), \ell = \ell_{focal}$. $f$ encodes the angle into a discrete distribution, and $\ell$ measures quality of classification; \textbf{2)} in the case of PSC \cite{yu2023psc}, $f=\cos(\theta+\varphi_i), i=1...N, \ell = \ell_{l1}$. $f$ encodes the angle into a continuous vector, $\ell$ measures the encoded vector distance.

Compared with the diverse optimization forms (independent or joint) for $f$, what is more noteworthy is encoding mode of $f$. Note that the model in \cref{eq:loss-smoothing} outputs $\theta$, $f$ is explicitly applied in loss calculation, while the model in \cref{eq:angle-smoothing} directly outputs the value encoded by $f$. For detector $F$, the former's fitting target is still $\theta_{box} \sim \theta_{obj}$ with a break point, while the latter's target becomes $f_{box} \sim f_{obj}$. Thanks to the periodic aggregation properties of $f$, the differences between box and object are eliminated, which will no longer suffer from difficulty about fitting breakpoints.

To summary up, it is a better choice to make model directly fit the smooth value rather than utilize it just in loss calculation. As for the reason why joint-optim methods do not adopt such design, it is most likely because it is difficult to recover the angle from the joint-encoding of the model output. Dramatically, the advantages of joint optimization outweigh the disadvantages of loss-smoothing, which eventually misleads researchers to believe that the boundary problem can be solved by joint optimization.

\begin{figure*}[t]
   \centering
   \includegraphics[width=0.85\linewidth]{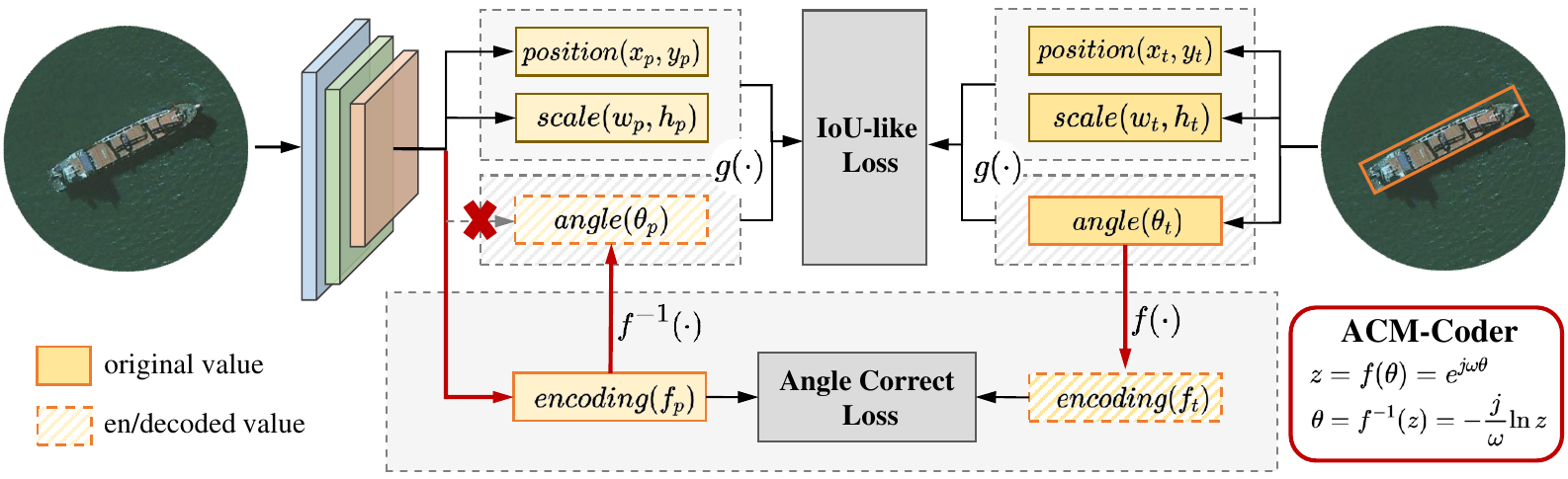}
   \caption{Overview of proposed Dual-Optimization paradigm and ACM-Coder. The detector outputs angular ACM-encoding $f_p$, subject to explicit supervision. On this basis, another IoU-like loss based on joint-encoded $g(\cdot)$ is applied onto ACM-decoded angle $f^{-1}(f_p)$. The paradigm achieves the objectives of both correcting angular boundary and blending parameters.}
   \label{fig:overview}
\end{figure*}

\section{Method}
\subsection{Dual-Optimization for Angle}
\label{sec:acm}
Considering that joint optimization has become the mainstream scheme at present, a convenient improvement strategy is to correct angle by angle-smoothing independent optimization with reversible coding, as well as to blend angle with other parameters by joint optimization based on corrected angle, which can be formulated as follows:
\begin{equation}
\begin{aligned}
\label{eq:mix-smoothing}
\theta 
= f^{-1} & \Bigl (\mathop{\arg\min}\limits_{z=f_p} \left [\ell_f + \ell_g \right ] \Bigr ) \\
s.t. \ &\ell_f = \ell \bigl (z; f(\theta_t) \bigr) \\
 \ & \ell_g = \ell \bigl (g(f^{-1}(z));g(\theta_t) \bigr )
\end{aligned}
\end{equation}
where model still fits continous $f_p$, and $f,g$ are encoding function in independent/joint optimization.

Since $f^{-1}$ participates in loss calculation in joint-optimization, $f$ not only needs to be continuous, differentiable, and reversible (GWD/KLD/FKIoU/SkewIoU fail to satisfy), but its inverse function $f^{-1}$ also needs to satisfy these properties. Discrete encodings like CSL rely on $\argmin$ to make $f^{-1}$ nondifferentiable, so only continuous encodings like PSC remains a chance to be differentiable.

To this end, we propose a \textbf{Angle Correct Module (ACM)} based on complex-exponential function. The module implements the above $f$ and $f^{-1}$, which can be easily plugged into the existing workflow of oriented object detectors to repair angular prediction. 
As is shown in \cref{fig:overview}, the detector needs to output angular encoding rather than angle itself when ACM works, since a consistent attribute ($f$, just similar to $x,y,w,h$) for both box and object can never cause boundary discontinity problem. This means $f(\theta_{box})=f(\theta_{obj})$, which is equivalent to $f(\theta_{obj} \mod \pi)=f(\theta_{obj})$ due to \cref{eq:sawtooth}, so $f(0)=f(\pi)$. To recover the unique angular value for box, $f$ needs to be reversible at least in $[0, \pi)$. This also means that $f$ is continuous over $[0, \pi]$, and it will causes a many-to-one correspondences not only at the interval boundary but also at other points according to Rolle's theorem. The contradiction implies it impossible to find a eligible $f$.

However, the "impossibility" mentioned above is only restricted to the most common case where $f$ belongs to real number domain $\mathbb{R}$. When we broaden our perspective to the complex number domain $\mathbb{C}$, the miracle will occur even without any bells and whistles. We will achieve the goal of reversible transformation by the simplest, yet the most classic complex transformation, i.e., complex-exponential transformation, as following:
\begin{align}
    z &= f (\theta) = e^{j\omega\theta} 
    \label{eq:acm_encode}\\
    \theta &= f^{-1}(z) = -\frac{j}{\omega} \ln z 
    \label{eq:acm_decode}
\end{align}    
where $z \in \mathbb{C}$ is encoded value, 
$j$ represents imaginary unit, and $\omega \in \mathbb{R}^+$ is angular frequency.
Due to \cref{eq:acm_decode} decoding can get unique angle only in a single cycle on the complex plane, $\omega\theta$ 's range $[0, \omega\pi) \subseteq [0, 2\pi)$, so it is necessary to satisfy $\omega \le 2$. To determine the appropriate $\omega$, we discuss the relationship of $f_{box} \sim f_{obj}$ as following:
\begin{equation}
\label{eq:f_box_obj}
\begin{aligned}
f_{box} 
&= e^{j\omega\theta_{box}}
= e^{j\omega(\theta_{obj} \bmod \pi)} \\
&=
\begin{cases}
\begin{aligned}
\begin{aligned}
&&& e^{j\omega \theta_{obj}}, &&\theta_{obj} \in [0, \pi) \\
&&& e^{j\omega \theta_{obj}} \cdot e^{-j\omega\pi}, && \theta_{obj} \in [\pi, 2\pi)
\end{aligned}
\end{aligned}
\end{cases}
\end{aligned}
\end{equation}

Through further derivation of the formula, we can find that \textbf{1)} When $\omega=2$, \cref{eq:f_box_obj} can be simplified to a straightforward $f_{box} = f_{obj}$. $f$ becomes a consistent attribute for both box and object, and it is perfectly in line with our design goals; \textbf{2)} When $\omega=1$, \cref{eq:f_box_obj} can be just simplified to a $f_{obj} \cdot \text{sign}(\pi - \theta_{obj})$. $f_{box}$ and $f_{obj}$ has a simple relationship but still with breakpoints; \textbf{3)} When $\omega \neq 2$ and $\omega \neq 1$, $e^{-j\omega\pi}$ is no longer a real factor, which makes \cref{eq:f_box_obj} difficult to simplify, and $f_{box} \sim f_{obj}$ difficult to analyze. To sum up, we finally choose $\omega=2$ in ACM. More details will be provided in the supplementary materials.

\subsection{Loss Functions}
As is shown in \cref{fig:overview}, given a batch of images, the detector outputs the classification $c_p$, position $(x_p, y_p)$, scale $(w_p, h_p)$, and angular encoding $f_{p}$, and the corresponding ground truth is $c_t$, $(x_t, y_t)$, $(w_t, h_t)$, and $\theta_t$.
First, we calculate the loss of the angular encoding in ACM, which is
\begin{equation}
\mathcal{L}_{acm} = \ell_{smooth\_l1} \bigl (f_p, f_t \bigr )
\end{equation}
Then, we jointly optimize the decoded angle $\theta_p=f_p^{-1}$,  with other parameters (abbreviated as $xywh$), which is
\begin{equation}
\mathcal{L}_{box} = \ell \bigl (B(xywh_p, \theta_p), B(xywh_t, \theta_t) \bigr )
\end{equation}
where $\ell \in \{\ell_{riou}, \ell_{kld}, \ell_{gwd}, ...\}$.
In addition, we also calculate the classification loss, which is
\begin{equation}
\mathcal{L}_{cls} = \ell_{focal} \bigl (c_p, c_t \bigr )    
\end{equation}
Finally, the total loss is as follows ($\lambda_{box}, \lambda_{acm}$ are coefficients to balance each parts of loss):
\begin{equation}
\mathcal{L} = \mathcal{L}_{cls} + \lambda_{box}\mathcal{L}_{box} + \lambda_{acm}\mathcal{L}_{acm}    
\end{equation}
By default, we set $\lambda_{box}=1, \lambda_{acm}=0.2$ in experiments.


\subsection{Differences With Other Coding Methods}
The difference between ACM and vanilla joint-optim encoding methods is self-evident, hence the focus here is primarily on independently-optim encoding methods, especially PSC \cite{yu2023psc}, which is a continuous encoding as ACM.
ACM and PSC are distinct under any circumstances. Basically, the encoding formula of PSC is $\{\cos(\omega \theta + \varphi_i) \ | i=1...N, N\geq3\}$, while ACM is $e^{j\omega \theta}=\cos(\omega \theta)+j\sin(\omega \theta)$. 
Obviously, ACM has a different theory and form. Moreover, they cannot be interconverted, even with PSC's $N$ at 4.
Specifically, when PSC's $N$ equals 4, the encoding vector becomes $\boldsymbol{f}=(-c, -s, c, s)$, where $c$ \& $s$ are abbreviation of $\cos$ \& $\sin$. Note that $\boldsymbol{f}$ is just ground-truth, its prediction $\boldsymbol{\hat{f}}$ has some deviation $\boldsymbol{e}$ compared with $\boldsymbol{f}$, i.e. $\boldsymbol{\hat{f}}=\boldsymbol{f}+\boldsymbol{e}$. But corresponding prediction-heads are independent modules without pairwise-constraints 
($e_1 + e_3 = 0$, $e_2 + e_4 = 0$), so $\boldsymbol{\hat{f}}$ is no longer ensured as two duplicate values $(c,s)$, i.e., $\hat{f}_1 \neq -\hat{f}_3, \hat{f}_2 \neq -\hat{f}_4$.
In contrast, complex encoding $e^{j\omega \theta}$ of ACM is equivalent to  vector $(c,s)$, where pairwise-constraints have been implied.
Consequently, the decoded angular prediction will naturally differ from ACM. \textbf{The stronger constraints/priors implied in ACM reduce optimization difficulty of entire model}, making ACM better.
\section{Experiment}

\begin{table}[tb!]
\caption{Ablation study of different encoding length.}
\centering
\resizebox{0.60\linewidth}{!}{
\begin{tabular}{c|c|cc}
\toprule
\multirow{2}{*}{\textbf{Method}} & \multirow{1}{*}{\textbf{Encoding}} & 
\multicolumn{2}{c}{\textbf{HRSC2016}}  \\
& \textbf{Length} & 
\textbf{AP$_{50}$}&  \textbf{AP$_{75}$} \\
\midrule
\multirow{1}{*}{Direct}
& 1 & 88.26 & 62.95 \\
\midrule
\multirow{4}{*}{CSL}
& 3 & 19.50 & 2.56 \\
& 60 & 48.96 & 13.58 \\
& 90 & 90.49 & 61.43 \\
& 180 & 90.53 & 77.76 \\
\midrule
\multirow{4}{*}{PSC}
& 3 & 89.91 & 79.20\\
& 20 & 90.55 & 79.54\\
& 60 & 90.62 & 79.86\\
& 180 & 90.56 & 79.51\\
\midrule
\rowcolor{gray!20} \multirow{1}{*}{ACM}
& 2 & 90.57 & 86.33\\
\bottomrule
\end{tabular}}
\label{tab:encoding_length}
\end{table}

\subsection{Datasets}

\textbf{DOTA\cite{xia2018dota}} is one of the largest datasets for oriented object detection in aerial images, which contains 2,806 images with fifteen categories of 188,282 instances in total. 
The training, validation and testing set include 1411, 458 and 937 images, respectively.
The categories are defined as: Plane (PL), Baseball Diamond (BD), Bridge (BR), Ground Field Track (GTF), Small Vehicle (SV), Large Vehicle (LV), Ship (SH), Tennis Court (TC), Basketball Court (BC), Storage Tank (ST), Soccer-Ball Field (SBF), Roundabout (RA), Harbor (HA), Swimming Pool (SP), and Helicopter (HC).
We crop training images into the patches of size $1024\times1024$ pixels with an overlap of 256 pixels.
When testing, we crop the testing set images into $4000\times4000$ patches with an overlap of 1024 pixels, to mitigate the negative impact of the cutting.

\noindent \textbf{HRSC2016\cite{liu2017high}} contains images from two scenarios with ships on sea and close inshore. The training, validation and testing set include 436, 181 and 444 images, with the image size ranging from 300 $\times$ 300 to 1500 $\times$ 900. 
We adjust the long side of each image to a fixed size (640 pixels) and keep the original aspect ratio for training and testing.

\noindent \textbf{UCAS-AOD\cite{zhu2015orientation}}  contains two categories: Car and Plane, which includes 1,510 aerial images of about 659 $\times$ 1,280 pixels, with two categories of 14,596 instances in total. 
We randomly select 1,110 for training and 400 for testing. 
Finally, we adopt the same data processing strategy as HRSC2016.

\subsection{Implementation Details}
\textbf{Evaluation Metric.} 
Methods are evaluated using the standard COCO style Average Precision (AP) \cite{lin2014microsoft}, which is the convention throughout the field of object detection. It is worth noting that AP$_{75}$ is gradually replacing AP$_{50}$ as the most reliable metric for oriented object detection due to AP$_{75}$'s higher sensitivity to angle deviation than AP$_{50}$. Following mainstream works \cite{yang2021learning,zeng2023ars}, we adopt AP$_{75}$ as main metric, while AP$_{50}$ is auxiliary metric. 

\begin{table}[t]
\caption{Ablation study of different encoding mode.}
\centering
\resizebox{0.85\linewidth}{!}{
\begin{tabular}{c|c|cc|cc}
\toprule
\multirow{2}{*}{\textbf{Method}} & \multirow{1}{*}{\textbf{Encoding}} & 
\multicolumn{2}{c|}{\textbf{HRSC2016}} & \multicolumn{2}{c}{\textbf{DOTA-v1.0}} \\
& \textbf{Mode}
& \textbf{AP$_{50}$}&  \textbf{AP$_{75}$} 
& \textbf{AP$_{50}$}&  \textbf{AP$_{75}$} \\
\midrule
\multirow{1}{*}{Direct}
& n/a & 88.26 & 62.95 & 71.97 & 26.11 \\
\midrule
\multirow{2}{*}{CSL}
& \cellcolor{gray!20}implicit & \cellcolor{gray!20}90.53 & \cellcolor{gray!20}77.76 & \cellcolor{gray!20}70.83 & \cellcolor{gray!20}38.71\\
& explicit & 6.06 & 1.05 & 33.29 & 10.90\\
\midrule
\multirow{2}{*}{PSC}
& \cellcolor{gray!20}implicit & \cellcolor{gray!20}89.91 & \cellcolor{gray!20}79.20 & \cellcolor{gray!20}71.41 & \cellcolor{gray!20}39.35\\
& explicit & 53.43 & 33.65 & 50.02 & 23.08\\
\midrule
\multirow{2}{*}{ACM}
& \cellcolor{gray!20}implicit & \cellcolor{gray!20}90.57 & \cellcolor{gray!20}86.33 & \cellcolor{gray!20}74.99 & \cellcolor{gray!20}41.44\\
& explicit & 54.66 & 31.45 & 50.67 & 19.91\\
\bottomrule
\end{tabular}}
\label{tab:encoding_position}
\end{table}

\begin{table}[tb!]
\caption{Ablation study of different supervision.}
\centering
\resizebox{0.85\linewidth}{!}{
\begin{tabular}{c|c|c|cc|cc}
\toprule
\textbf{Model} & 
\multirow{2}{*}{$\boldsymbol{\mathcal{L}_{acm}}$} & \multirow{2}{*}{$\boldsymbol{\mathcal{L}_{box}}$} & \multicolumn{2}{c|}{\textbf{HRSC2016}} & \multicolumn{2}{c}{\textbf{DOTA}} \\
\textbf{Output} & & & \textbf{AP$_{50}$} & \textbf{AP$_{75}$} & \textbf{AP$_{50}$} & \textbf{AP$_{75}$} \\
\midrule
$\theta$ & n/a & n/a & 88.26 & 62.95 & 71.97 & 26.11 \\
\midrule
\multirow{3}{*}{$f_p$} & $\checkmark$ & & 
90.57 & 86.33  & 74.99 & 41.44  \\
&  & $\checkmark$ & 
37.37 & 13.98  & 54.67 & 19.67  \\
& $\cellcolor{gray!20}\checkmark$ & \cellcolor{gray!20}$\checkmark$ & 
\cellcolor{gray!20}90.47 & \cellcolor{gray!20}88.33 & \cellcolor{gray!20}74.21 & \cellcolor{gray!20}42.83  \\
\bottomrule
\end{tabular}}
\label{tab:different_supervision}
\end{table}

\noindent \textbf{Training Details.}
All approaches are implemented in PyTorch, and training is done on NVIDIA RTX 3090 GPUs. We choose the anchor-free method CenterNet \cite{zhou2019objects} to build the rotated detector and ImageNet pretrained ResNet-50 \cite{he2016deep} as the backbone. The network is optimized by Adam for 140 epochs with the learning rate dropped by 10$\times$ at 100 and 130 epochs. 
As the DOTA dataset takes a large image resolution as an input, the batch size is set as 12 with an initial learning rate $1.25\times 10^{-4}$.
For the HRSC2016 and UCAS-AOD datasets, the batch size is set as 32, and the initial learning rates are set as $2\times$ $10^{-4}$ and $1\times$ $10^{-4}$, respectively. 


\begin{table*}[tb!]
\caption{Ablation study on various datasets with different joint-optim loss. Base detector is CenterNet.}
\label{tab:hrsc_ucas_dota}
\centering
\resizebox{0.93\textwidth}{!}{
\begin{tabular}{l|cc|cc|cc|cc}
\toprule
\multirow{2}{*}{\textbf{Method}} & \multicolumn{2}{c|}{\textbf{HRSC2016 (Ship)}} & \multicolumn{2}{c|}{\textbf{UCAS-AOD (Car)}} & \multicolumn{2}{c|}{\textbf{UCAS-AOD (Plane)}}& \multicolumn{2}{c}{\textbf{DOTA-v1.0}}\\
& \textbf{AP$_{50}$}&  \textbf{AP$_{75}$} 
& \textbf{AP$_{50}$}&  \textbf{AP$_{75}$} 
& \textbf{AP$_{50}$}&  \textbf{AP$_{75}$} 
& \textbf{AP$_{50}$}&  \textbf{AP$_{75}$}\\
\midrule
GWD \cite{yang2021rethinking} & 84.94 & 61.87 & 87.25 & 28.46 & 90.34 & 38.22  &  73.12 & 34.98\\
\rowcolor{gray!20} ACM-GWD  &  90.63 \textbf{\textcolor[rgb]{0,0.6,0}{\small (+5.69)}} &  86.71 \textbf{\textcolor[rgb]{0,0.6,0}{\small (+24.84)}} &  88.69 \textbf{\textcolor[rgb]{0,0.6,0}{\small (+1.44)}} &  29.15 \textbf{\textcolor[rgb]{0,0.6,0}{\small (+0.69)}} &  90.35 \textbf{\textcolor[rgb]{0,0.6,0}{\small (+0.01)}} &  76.00 \textbf{\textcolor[rgb]{0,0.6,0}{\small (+37.78)}} &  73.71 \textbf{\textcolor[rgb]{0,0.6,0}{\small (+0.59)}}  &  41.97 \textbf{\textcolor[rgb]{0,0.6,0}{\small (+6.99)}}\\
\midrule
KLD \cite{yang2021learning} &  90.01 &79.29 & 87.54 & 29.99  &90.33 & 29.19 & 73.41 & 35.25\\
\rowcolor{gray!20} ACM-KLD &   90.55 \textbf{\textcolor[rgb]{0,0.6,0}{\small (+0.54)}} &  87.45 \textbf{\textcolor[rgb]{0,0.6,0}{\small (+8.16)}} &  88.76 \textbf{\textcolor[rgb]{0,0.6,0}{\small (+1.22)}} &  30.40 \textbf{\textcolor[rgb]{0,0.6,0}{\small (+0.41)}} &  90.39 \textbf{\textcolor[rgb]{0,0.6,0}{\small (+0.06)}} &  75.65 \textbf{\textcolor[rgb]{0,0.6,0}{\small (+46.46)}} &  73.95 \textbf{\textcolor[rgb]{0,0.6,0}{\small (+0.54)}} &  42.97 \textbf{\textcolor[rgb]{0,0.6,0}{\small (+7.72)}}\\
\midrule
KFIoU \cite{yang2023kfiou} & 88.26 & 62.95 &85.74 & 24.44  &90.34 & 16.81 &  71.97 & 26.11\\
\rowcolor{gray!20} ACM-KFIoU &   90.55 \textbf{\textcolor[rgb]{0,0.6,0}{\small (+2.29)}} &  87.77 \textbf{\textcolor[rgb]{0,0.6,0}{\small (+24.82)}} &  88.31 \textbf{\textcolor[rgb]{0,0.6,0}{\small (+2.57)}} &  34.81 \textbf{\textcolor[rgb]{0,0.6,0}{\small (+10.37)}} &  90.40 \textbf{\textcolor[rgb]{0,0.6,0}{\small (+0.06)}} &  74.48 \textbf{\textcolor[rgb]{0,0.6,0}{\small (+57.67)}} &  74.51 \textbf{\textcolor[rgb]{0,0.6,0}{\small (+2.54)}} &  40.49 \textbf{\textcolor[rgb]{0,0.6,0}{\small (+14.38)}}\\
\midrule
SkewIoU \cite{zhou2019iou} &  89.39 & 76.43 &87.73 & 27.59  &90.34 & 63.64 &  {73.62} & {38.01}\\
\rowcolor{gray!20} ACM-SkewIoU &  90.47 \textbf{\textcolor[rgb]{0,0.6,0}{\small (+1.08)}} &  88.33 \textbf{\textcolor[rgb]{0,0.6,0}{\small (+11.09)}} &  88.27 \textbf{\textcolor[rgb]{0,0.6,0}{\small (+0.54)}} &  29.13 \textbf{\textcolor[rgb]{0,0.6,0}{\small (+1.74)}} &  90.37 \textbf{\textcolor[rgb]{0,0.6,0}{\small (+0.03)}} &  75.13 \textbf{\textcolor[rgb]{0,0.6,0}{\small (+11.49)}} & 74.21 \textbf{\textcolor[rgb]{0,0.6,0}{\small (+0.59)}} &  42.83 \textbf{\textcolor[rgb]{0,0.6,0}{\small (+4.37)}}\\
\bottomrule
\end{tabular}}
\end{table*}

\begin{figure*}[!tb]
   \centering
   \includegraphics[width=0.93\linewidth]{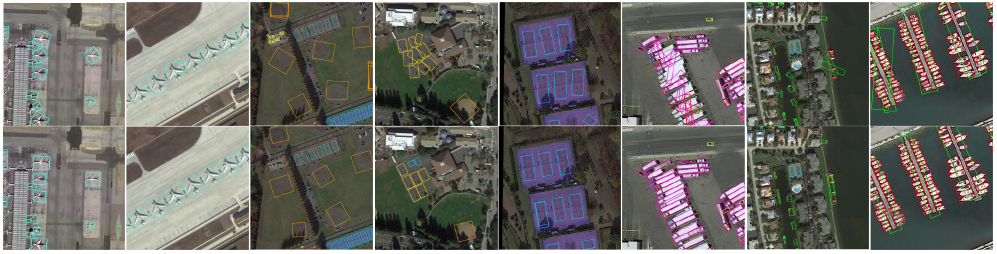}
   \caption{Visualized comparison of detection results between KFIoU \cite{yang2023kfiou} and enhanced ACM-KFIoU. The images are arranged from left to right in order of increasing aspect-ratio of objects, and the first-col and bottom-col are the results of KFIoU and ACM-KFIoU, respectively. Our ACM greatly eliminates the angular prediction errors in the original KFIoU.}
   \label{fig:vis_hrsc_ucas}
\end{figure*}

\subsection{Ablation Studies}

\noindent \textbf{Different encoding length.} 
To comprehensively compare the costs and performance of various encoding methods, we conducted experiments as \cref{tab:encoding_length}. The results indicate that CSL relies on relatively long encoding, while the choice of coding length for PSC is challenging.
Compared with these methods, ACM achieves the best performance without the need to choose the encoding length, which is fixed to $2$.

\noindent \textbf{Different encoding mode.} 
To validate our motivation, that is, implicit encoding is superior to explicit encoding, we conducted comparative experiments based on various encoding methods as \cref{tab:encoding_position}. The results show that, even for the identical encoding methods, performance is notably weaker when predicting the angle itself \textit{(explicit)} rather than  angular encoding \textit{(implicit)}. This is an observation that has long been overlooked, yet it points towards a viable direction for addressing boundary discontinuity problem in the future.

\noindent \textbf{Different supervision.}
To validate the necessity of each supervision of proposed dual-optimization, we conducted experiments as \cref{tab:different_supervision}. The results show that, compared with full dual-optimization, performance decreased (yet still surpass baseline) after the removal of refine supervision, but there was a severe decline in performance after the removal of correct supervision. 
We think it is difficult to ensure that predicted frequencies align with the preset decoded frequencies if without explicit supervision, and such mismatch will make decoded angles fall into a wrong range, so performance degenerates. By the way, since the size of DOTA dataset is much larger than HRSC2016 dataset, the mismatch can get some mitigation, but just like a drop in the ocean.

\begin{table*}[!tb]
\caption{Performance of different detectors on DOTA-v1.0 dataset. MS indicates that multi-scale training/testing is used.} 
\centering
\resizebox{1.0\linewidth}{!}{
\begin{tabular}{l|c|c|c|c|c|c|c|c|c|c|c|c|c|c|c|c|c} 
\toprule
\multicolumn{1}{c|}{\textbf{Method}} & MS& \textbf{PL} & \textbf{BD} & \textbf{BR} & \textbf{GTF} & \textbf{SV} & \textbf{LV} & \textbf{SH} & \textbf{TC} & \textbf{BC} & \textbf{ST} & \textbf{SBF} & \textbf{RA} & \textbf{HA} & \textbf{SP} & \textbf{HC} & \textbf{AP$_{50}$} \\ 
\midrule
PIoU~\cite{chen2020piou}&  & 80.90 & 69.70 & 24.10 & 60.20 & 38.30 & 64.40 & 64.80 & 90.90 & 77.20 & 70.40 & 46.50 & 37.10 & 57.10 & 61.90 & 64.00 & 60.50  \\
RoI-Trans.~\cite{ding2019learning}&\checkmark   & 88.64 & 78.52 & 43.44 & 75.92 & 68.81 & 73.68 & 83.59 & 90.74 & 77.27 & 81.46 & 58.39 & 53.54 & 62.83 & 58.93 & 47.67 & 69.56  \\
O$^2$-DNet~\cite{wei2020oriented}&\checkmark   & 89.31 & 82.14 & 47.33 & 61.21 & 71.32 & 74.03 & 78.62 & 90.76 & 82.23 & 81.36 & 60.93 & 60.17 & 58.21 & 66.98 & 61.03 & 71.04  \\
DAL~\cite{ming2021dynamic}&\checkmark & 88.61 & 79.69 & 46.27 & 70.37 & 65.89 & 76.10 & 78.53 & 90.84 & 79.98 & 78.41 & 58.71 & 62.02 & 69.23 & 71.32 & 60.65 & 71.78  \\
P-RSDet~\cite{zhou2020arbitrary}&\checkmark   & 88.58 & 77.83 & 50.44 & 69.29 & 71.10 & 75.79 & 78.66 & 90.88 & 80.10 & 81.71 & 57.92 & 63.03 & 66.30 & 69.77 & 63.13 & 72.30  \\
BBAVectors~\cite{yi2021oriented}&\checkmark  & 88.35 & 79.96 & 50.69 & 62.18 & 78.43 & 78.98 & 87.94 & 90.85 & 83.58 & 84.35 & 54.13 & 60.24 & 65.22 & 64.28 & 55.70 & 72.32  \\
 DRN~\cite{pan2020dynamic}&\checkmark    & 89.71 & 82.34 & 47.22 & 64.10 & 76.22 & 74.43 & 85.84 & 90.57 & 86.18 & 84.89 & 57.65 & 61.93 & 69.30 & 69.63 & 58.48 & 73.23  \\
CFC-Net~\cite{ming2021cfc}&\checkmark    & 89.08 & 80.41 & 52.41 & 70.02 & 76.28 & 78.11 & 87.21 & 90.89 & 84.47 & 85.64 & 60.51 & 61.52 & 67.82 & 68.02 & 50.09 & 73.50  \\
Gliding Vertex~\cite{xu2020gliding}&    & 89.64 & 85.00 & 52.26 & 77.34 & 73.01 & 73.14 & 86.82 & 90.74 & 79.02 & 86.81 & 59.55 & 70.91 & 72.94 & 70.86 & 57.32 & 75.02  \\
Mask OBB~\cite{wang2019mask}&\checkmark     & 89.56 & 85.95 & 54.21 & 72.90 & 76.52 & 74.16 & 85.63 & 89.85 & 83.81 & 86.48 & 54.89 & 69.64 & 73.94 & 69.06 & 63.32 & 75.33   \\
CenterMap~\cite{wang2020learning}&\checkmark    & 89.83 & 84.41 & 54.60 & 70.25 & 77.66 & 78.32 & 87.19 & 90.66 & 84.89 & 85.27 & 56.46 & 69.23 & 74.13 & 71.56 & 66.06 & 76.03  \\
CSL~\cite{yang2020arbitrary}&\checkmark    & 90.25 & 85.53 & 54.64 & 75.31 & 70.44 & 73.51 & 77.62 & 90.84 & 86.15 & 86.69 & 69.60 & 68.04 & 73.83 & 71.10 & 68.93 & 76.17  \\ 
R$^3$Det~\cite{yang2021r3det}&\checkmark   & 89.80 & 83.77 & 48.11 & 66.77 & 78.76 & 83.27 & 87.84 & 90.82 & 85.38 & 85.51 & 65.67 & 62.68 & 67.53 & 78.56 & 72.62 & 76.47  \\
GWD~\cite{yang2021rethinking}&\checkmark   &  86.96 & 83.88 & 54.36 & 77.53 & 74.41 & 68.48 &	80.34 &	86.62 &	83.41 &	85.55 &	73.47 &	67.77 &	72.57 &	75.76 & 73.40 & 76.30  \\
SCRDet++~\cite{yang2022scrdet++}&\checkmark   & 90.05 & 84.39 & 55.44 & 73.99 & 77.54 & 71.11 & 86.05 & 90.67 & 87.32 & 87.08 & 69.62 & 68.90 & 73.74 & 71.29 & 65.08 & 76.81  \\
KFIoU~\cite{yang2023kfiou}&\checkmark   & 89.46 & 85.72 & 54.94 & 80.37 & 77.16 & 69.23 & 80.90 & 90.79 & 87.79 & 86.13 & 73.32 & 68.11 & 75.23 & 71.61 & 69.49 & 77.35  \\
DCL~\cite{yang2021dense}&\checkmark    & 89.26 & 83.60 & 53.54 & 72.76 & 79.04 & 82.56 & 87.31 & 90.67 & 86.59 & 86.98 & 67.49 & 66.88 & 73.29 & 70.56 & 69.99 & 77.37  \\
RIDet~\cite{ming2021optimization}&\checkmark   & 89.31 & 80.77 & 54.07  & 76.38   & 79.81  & 81.99 & 89.13 & 90.72  & 83.58   & 87.22  & 64.42  & 67.56  & 78.08  & 79.17 & 62.07  & 77.62   \\
PSC~\cite{yu2023psc}&\checkmark& 89.86 & 86.02 & 54.94 & 62.02 & 81.90 & 85.48 & 88.39 & 90.73 & 86.90 & 88.82 & 63.94 & 69.19 & 76.84 & 82.75 & 63.24 & 78.07  \\
KLD~\cite{yang2021learning}&\checkmark &88.91 &85.23 &53.64 &81.23 &78.20 &76.99 &84.58 &89.50 &86.84 &86.38 &71.69 &68.06 &75.95 &72.23 &75.42 &78.32 \\
\midrule
\rowcolor{gray!20} CenterNet-ACM&\checkmark & 89.84 & 85.50 & 53.84 & 74.78 & 80.77 & 82.81 & 88.92 & 90.82 & 87.18 & 86.53 & 64.09 & 66.27 & 77.51 & 79.62 & 69.57 & \textbf{78.53} \\ 
\rowcolor{gray!20} RoI-Trans.-ACM &\checkmark &85.55 &80.53 &61.21 &75.40 &80.35 &85.60& 88.32 &89.88 &87.13& 87.10& 68.15& 67.94& 78.75& 79.82& 75.96& \textbf{79.45} \\
\bottomrule
\end{tabular}}
\label{tab:dota}
\end{table*}

\noindent \textbf{Different joint-optim loss on various datasets.}
To eliminate the influence of the classification branch on the detection results, we conducted experiments on the datasets comprising simple scenes with only single-category objects per image.
The HRSC2016 dataset contains large aspect ratio ships, and the UCAS-AOD dataset contains rectangle cars and square-like planes. 
As is shown in \cref{tab:hrsc_ucas_dota}, both AP$_{50}$ and AP$_{75}$ get significant improvement from ACM on HRSC2016(Ship) and UCAS-AOD(Car). 

It is worth noting that the improvement of AP$_{50}$ are negligible on the UCAS-AOD(Plane) dataset, while the improvement of AP$_{75}$ are tremendous. 
It is never an accident, and the reasons include: 
\textbf{1)} For square-like objects, the IoU is always over 0.5 regardless of the angle of the predicted box, making AP$_{50}$ insensitive to square-like objects.
\textbf{2)} When a square-like box is converted to a 2D-Gaussian distribution, the 2D-Gaussian distribution is completely symmetric like a circle, which makes it impossible for these methods (GWD, KLD, KFIoU) based on 2D-Gaussian distribution to accurately predict the angle of square-like objects.
Since our ACM is friendly to square-like objects, it greatly improves these baseline methods based on 2D-Gaussian distribution by \textbf{37.78}\% (GWD), \textbf{46.46}\% (KLD) and \textbf{57.56}\% (KFIoU) on AP$_{75}$ on UCAS-AOD(Plane) dataset.

To explore performance in a more general cases, we conducted experiments on a dataset comprising complex scenes with only single-category objects in each image.
DOTA dataset contains a considerable number of categories and diverse environments. Experimental results at \cref{tab:hrsc_ucas_dota} show that the performance of all IoU-like methods are improved by \textbf{6.99}\% (GWD), \textbf{7.72}\% (KLD), \textbf{14.34}\% (KFIoU) and \textbf{14.34}\% (SkewIoU) on AP$_{75}$ after the ACM module is used. We also unexpectedly find that after the ACM module enhancement, both Gaussian-based loss and SkewIoU loss become very close in terms of AP$_{50}$ (\textbf{73.71}\%, \textbf{73.95}\%, \textbf{74.51}\%, \textbf{74.21}\%) and AP$_{75}$ (\textbf{41.97}\%, \textbf{42.97}\%, \textbf{40.49}\%, \textbf{42.83}\%), indicating that the primary distinction between them lies in their optimization capabilities for the angle.

\noindent \textbf{Visualized results.} 
We provide some visualization results on the DOTAv1.0 dataset as \cref{fig:vis_hrsc_ucas}. From detection results obtained by the KFIoU-based detector, we select some cases of poor angular prediction. Note that there exists slight angular deviations for boxes in KFIoU results sometimes, and significant angular errors in other times.
Fortunately, most angular errors are corrected in the results of ACM-KFIoU. It is also worth noting that ACM addresses the square-like-object case where KFIoU based on 2D Gaussian distribution fails for angular prediction. It is consistent with the quantitative results in \cref{tab:hrsc_ucas_dota}, and further verifies the effectiveness of our methods.

\subsection{Comparison with the State-of-the-Art}

\cref{tab:dota} presents a comprehensive comparison of recent detectors on DOTA-v1.0 dataset.  It is important to note that the performance of different methods may vary due to several factors, including image resolution, network architecture, detection framework, training strategies, and various optimization techniques employed.
In light of these variations, it becomes challenging to establish completely fair comparisons among the different approaches.  However, despite these challenges, our method has managed to achieve competitive results,  at around \textbf{78.53}\% / \textbf{79.45}\% on AP$_{50}$.  

\section{Conclusion}
In this paper, we experimentally find that widely used IoU-like methods do not actually solve the well-known boundary discontinuity problem. On further analysis, we find that the key to solution lies in the encoding mode of the smoothing function rather than in joint or independent optimization. 
Moreover, we propose a dual-optimization paradigm integrated with complex-exponential angular coding, which achieves the objectives of both correcting angular boundary and blending parameters. 
Finally, extensive experiments show that our methods effectively eliminate boundary problem and significantly improve detection performance for the object detector.

\section*{Acknowledgments}
This work is supported by National Key R\&D Program of China (2022YFD2001601), National Natural Science Foundation of China (62372433, 62072438, 61931008, U21B2024, 62071415), Zhejiang Provincial Natural Science Foundation of China(LDT23F01011F01, LDT23F01015F01, LDT23F01014F01), "Pioneer" and "Leading Goose" R\&D Program of Zhejiang Province (2022C01068).

{
    \small
    \bibliographystyle{ieeenat_fullname}
    \bibliography{main}
}


\maketitlesupplementary

\section{Another understanding of ACM}
\subsection{From Complex Function to Polar Mapping}

For the complex-exponential-based encoding proposed in the main text, leveraging Euler's formula allows for its transformation into a polar-coordinate mapping, as following:
\begin{align}
    z &= f (\theta) = e^{j\omega\theta} 
    \notag \\
    &=\cos(\omega\theta) + j \sin(\omega\theta)
    \\
    \theta &= f^{-1}(z) = -\frac{j}{\omega} \ln z 
    \notag \\
    &= \frac{1}{\omega}((\mathrm{arctan2}(z_{im}, z_{re}) + 2\pi) \bmod 2\pi) 
\end{align} 
where $\omega \in \mathbb{R}^+$ is angular frequency, $\mathrm{arctan2}$ is another version of $\mathrm{arctan}$ with quadrant assignment, and $z_{re}, z_{im}$ are real-part and imagine-part of complex coding $z \in \mathbb{Z}$, respectively. By hiding the complex mark of the encoding, we can regard it as a 2D polar coordinate encoding, as following:
\begin{align}
(z_x, z_y) &= f(\theta) = (\cos(\omega\theta), \sin(\omega\theta))
\\
\theta &= f^{-1}(z_x, z_y)
\notag \\
&=\frac{1}{\omega}((\mathrm{arctan2}(z_y, z_x) + 2\pi) \bmod 2\pi)
\end{align} 
where $\omega \in \mathbb{R}^+$ is still angular frequency, $\mathrm{arctan2}$ is another version of $\mathrm{arctan}$ with quadrant assignment, $z_x, z_y$ are x-axis-component and y-axis-component of 2D vector $\boldsymbol{z} \in \mathbb{R}^2$, respectively. This form is similar to continuous PSC encoding \cite{yu2023psc}, but note that PSC cannot perform the above conversion.

\subsection{Mathematical Meaning of Polar Mapping}
\begin{figure}[!tb]
\centering
\begin{minipage}[b]{0.4\linewidth}
  \centering
  \begin{subfigure}[b]{\linewidth}
    \includegraphics[width=\linewidth]{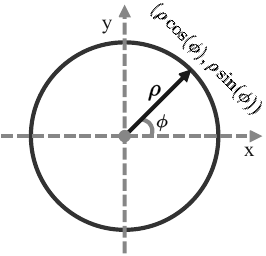}
    \caption{}
    \label{fig:poloar}
  \end{subfigure}
\end{minipage}
\hspace{5pt}
\begin{minipage}[b]{0.4\linewidth}
  \centering
  \begin{subfigure}[b]{\linewidth}
    \includegraphics[width=\linewidth]{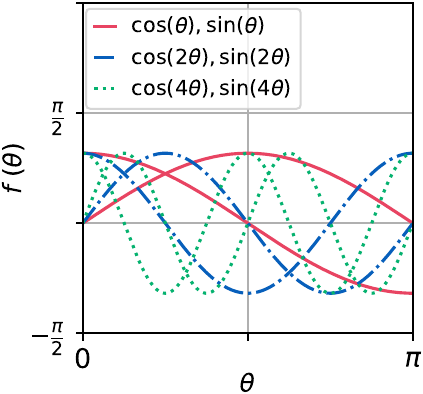}
    \caption{}
    \label{fig:f_theta}
  \end{subfigure}
\end{minipage}
\caption{Based on (a) polar coordinate decomposition, we define (b) a 2-dimensional wrapping function $f(\theta)$.}
\end{figure}

In this perspective, encoding corresponds to the Cartesian coordinates of a unit vector, while decoding corresponds to the polar coordinates representation of the same unit vector. As is shown in \cref{fig:poloar}, given a vector with polar angle $\phi$ and radius(length) $\rho$ in 2-dimensional space, it can be decompose as $(\rho\cos(\phi), \rho\sin(\phi))$ in Cartesian coordinates. When the radius $\rho$ is fixed and $\phi$ is just considered in single period, the polar angle and Cartesian coordinates are one-to-one correspondences. Therefore, even leaving aside we can utilize this relationship to design $f$ and obtain the corresponding $f^{-1}$ as \& \cref{fig:f_theta}. In contrast, PSC coding \cite{yu2023psc} does not have such a clear mathematical meaning, so it needs to be experimentally determined to encoding length hyperparameters.

\begin{table}
\caption{Ablation study of different encoding mode.}
\centering
\resizebox{0.95\linewidth}{!}{
\begin{tabular}{c|c|c|cc|cc}

			\toprule
			\multirow{2}{*}{$\omega=1$} & \multirow{2}{*}{$\omega=2$} & \multirow{2}{*}{$\omega=4$}  &\multicolumn{2}{c|}{\textbf{HRSC2016}} & \multicolumn{2}{c}{\textbf{DOTA}} \\
			\cline{4-7}
			 & & & \textbf{AP$_{50}$}&  \textbf{AP$_{75}$} & \textbf{AP$_{50}$}&  \textbf{AP$_{75}$} \\
			\cline{1-7}
                &   &     & 88.26 & 62.95 & 71.97 & 26.11  \\
			$\checkmark$ &   &     & 90.44 \textbf{\textcolor[rgb]{0,0.6,0}{\small (+2.18)}} & 78.90 \textbf{\textcolor[rgb]{0,0.6,0}{\small (+15.95)}} & 73.51 \textbf{\textcolor[rgb]{0,0.6,0}{\small (+1.54)}} & 39.29 \textbf{\textcolor[rgb]{0,0.6,0}{\small (+13.18)}}  \\
			    &  $\checkmark$  &   & \textbf{90.58} \textbf{\textcolor[rgb]{0,0.6,0}{\small (+2.32)}} & 86.12 \textbf{\textcolor[rgb]{0,0.6,0}{\small (+23.17)}} & 73.08 \textbf{\textcolor[rgb]{0,0.6,0}{\small (+1.11)}} & 39.62 \textbf{\textcolor[rgb]{0,0.6,0}{\small (+13.51)}} \\
			    &   & $\checkmark$ & 24.90 \textbf{\textcolor[rgb]{0.6,0,0}{\small (-63.36)}} & 20.82 \textbf{\textcolor[rgb]{0.6,0,0}{\small (-42.13)}} & 35.50 \textbf{\textcolor[rgb]{0.6,0,0}{\small (-36.47)}} & 17.29 \textbf{\textcolor[rgb]{0.6,0,0}{\small (-8.82)}} \\
			 \ours  & \ours $\checkmark$  & \ours $\checkmark$ & \ours 90.55 \textbf{\textcolor[rgb]{0,0.6,0}{\small (+2.29)}} & \ours \textbf{87.77} \textbf{\textcolor[rgb]{0,0.6,0}{\small (+24.82)}} & \ours \textbf{74.51} \textbf{\textcolor[rgb]{0,0.6,0}{\small (+2.54)}} & \ours \textbf{40.49} \textbf{\textcolor[rgb]{0,0.6,0}{\small (+14.38)}} \\
			\bottomrule
	\end{tabular}}
\label{tab:freq}
\end{table}

\begin{figure*}[t]
    \centering
    \begin{minipage}[b]{0.13\textwidth}
        \centering
        \begin{subfigure}[b]{\linewidth}
            \includegraphics[width=\textwidth]{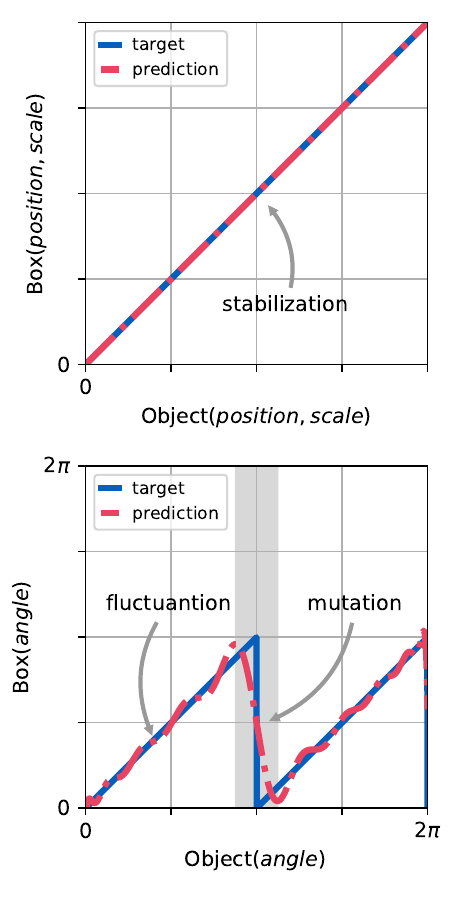}
            \caption{}
            \label{fig:original_wave}
        \end{subfigure}
    \end{minipage}
    \hspace{0.2cm}
    \begin{minipage}[b]{0.78\textwidth}
        \centering
        \begin{subfigure}[b]{\linewidth}
            \includegraphics[width=\textwidth]{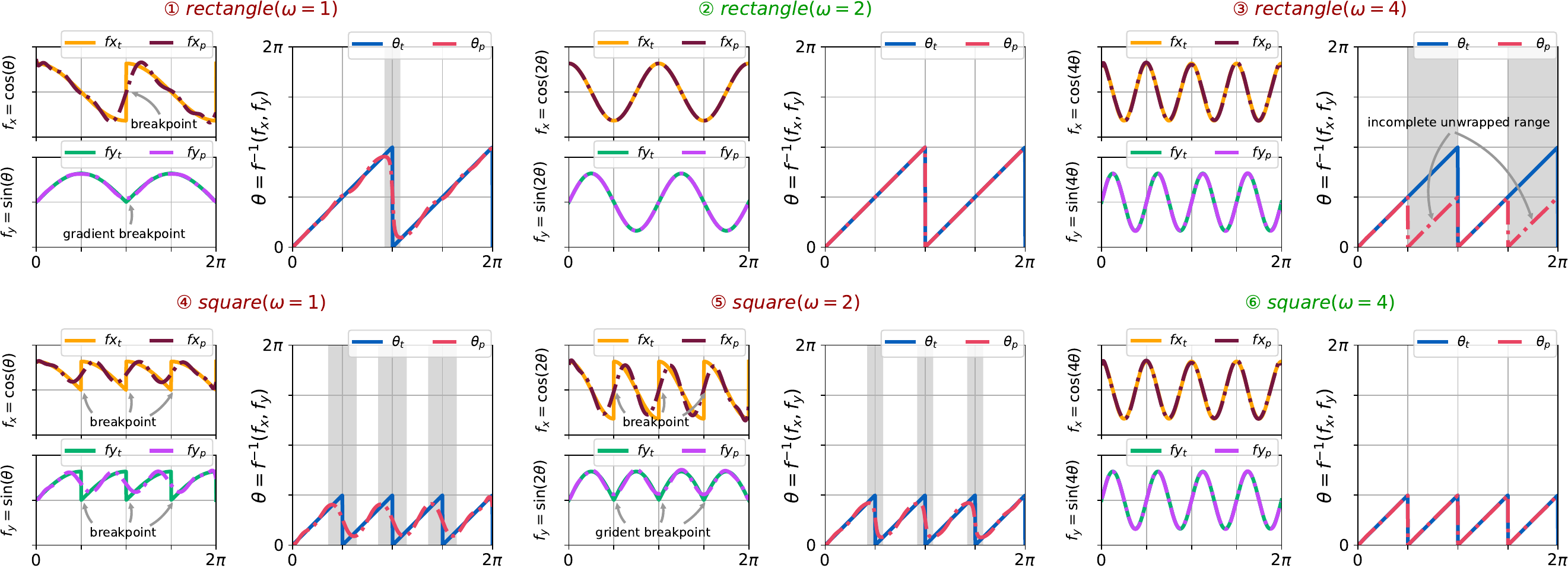}
            \caption{}
            \label{fig:sup_waves}
        \end{subfigure}
    \end{minipage}
    \caption{Waveform analysis: (a) original angular relationship between box and object. (b) based on polar-coordinate mapping, we define a 2-dimensional en/decoding function $f(\cdot)$. It will compound onto the original sawtooth wave $\theta_{box} = \theta_{obj} \pmod \pi$, and exhibits different effect for rectangular(top, $T=\pi$) \& square-like(bottom, $T=\frac{\pi}{2}$) objects when angular frequency $\omega=1,2,4$. The target and prediction are marked as solid line and dash line, respectively. Areas of inaccurate angular prediction are highlighted in gray. The optimal angular frequency for rectangular \& square-like objects is $2$ and $4$, respectively.}
\end{figure*}

\section{Determination of Angular Frequency}
\subsection{Perspective 1: Complex Function}
To determine the appropriate $\omega$, we discuss the relationship of $f_{box} \sim f_{obj}$ as following:
\begin{equation}
\label{eq:sup_f_box_obj}
\begin{aligned}
f_{box} 
&= e^{i\omega\theta_{box}}
= e^{i\omega(\theta_{obj} \bmod \pi)} \\
&=
\begin{cases}
\begin{aligned}
\begin{aligned}
&&& e^{i\omega \theta_{obj}}, &&\theta_{obj} \in [0, \pi) \\
&&& e^{i\omega \theta_{obj}} \cdot e^{-i\omega\pi}, && \theta_{obj} \in [\pi, 2\pi)
\end{aligned}
\end{aligned}
\end{cases}
\end{aligned}
\end{equation}

\textbf{1)} When $\omega=2$, $e^{-i\omega\pi}=1$, then
\begin{equation}
\begin{aligned}
f_{box}
&= e^{i\omega\theta_{box}}\\
&=
\begin{cases}
\begin{aligned}
\begin{aligned}
&&& e^{i\omega \theta_{obj}}, &&\theta_{obj} \in [0, \pi) \\
&&& e^{i\omega \theta_{obj}}, && \theta_{obj} \in [\pi, 2\pi)
\end{aligned}
\end{aligned}
\end{cases}\\
&=
f_{obj}
\end{aligned}
\end{equation}

\textbf{2)} When $\omega=1$, $e^{-i\omega\pi}=-1$, then
\begin{equation}
\begin{aligned}
f_{box}
&= e^{i\omega\theta_{box}}\\
&=
\begin{cases}
\begin{aligned}
\begin{aligned}
&&& e^{i\omega \theta_{obj}}, &&\theta_{obj} \in [0, \pi) \\
- &&& e^{i\omega \theta_{obj}}, && \theta_{obj} \in [\pi, 2\pi)
\end{aligned}
\end{aligned}
\end{cases}\\
&=
\begin{cases}
\begin{aligned}
\begin{aligned}
&&& f_{obj}, &&\theta_{obj} \in [0, \pi) \\
- &&& f_{obj}, && \theta_{obj} \in [\pi, 2\pi)
\end{aligned}
\end{aligned}
\end{cases}\\
&=
f_{obj} \cdot \text{sign}(\pi - \theta_{obj})
\end{aligned}
\end{equation}

Through further derivation of the formula, we can find that \textbf{1)} When $\omega=2$, \cref{eq:sup_f_box_obj} can be simplified to a straightforward $f_{box} = f_{obj}$. $f$ becomes a consistent attribute for both box and object, and it is perfectly in line with our design goals; \textbf{2)} When $\omega=1$, \cref{eq:sup_f_box_obj} can be just simplified to a $f_{obj} \cdot \text{sign}(\pi - \theta_{obj})$. $f_{box}$ and $f_{obj}$ has a simple relationship but still with breakpoints; \textbf{3)} When $\omega \neq 2$ and $\omega \neq 1$, $e^{-i\omega\pi}$ is no longer a real factor, which makes \cref{eq:sup_f_box_obj} difficult to simplify, and $f_{box} \sim f_{obj}$ difficult to analyze. To sum up, we finally choose $\omega=2$ in ACM.

\subsection{Perspective 2: Polar Mapping}
From the perspective of polar-coordinate mapping, ACM has more clear mathematical meaning and simple real number expression, so that we can carry out more direct analysis.
Although the single dimensional $\cos(\omega\theta)$ and $\sin(\omega\theta)$ are many-to-one, integration of them can achieve a one-to-one effect in a higher dimension, making $f$ an reversible transformation.
Due to polar coordinate decoding can get unique angle only in a single cycle, $\omega\theta$ 's range $[0, \omega\pi) \subseteq [0, 2\pi)$, so it is necessary to satisfy $\omega \le 2$. 

With encoding operation, original relationship $\theta_{box} \sim \theta_{obj}$ (\cref{fig:original_wave}) becomes $f_{box} \sim \theta_{obj}$ (\cref{fig:sup_waves}), where $f_{box}$ is the result of  encoding function $f(\cdot)$ applied on $\theta_{box}$. Thus, the waveform of $f_{box} \sim \theta_{obj}$ at $[0, 2\pi)$ is equivalent to repeating the encoded $\sin / \cos$ waveform at $[0, \pi)$ twice, due to the sawtooth wave of $\theta_{box} \sim \theta_{obj}$.

Obviously, the main issue of $\omega > 2$ (e.g. $\omega > 4$) lies in the incomplete decoding range, which will have a serious impact on angular prediction. In the valid angular frequency range, only when $\omega=2$, both encoding components are smooth (continuous and with continuous gradient) at $\theta_{obj} = \pi$. It indeed completely eliminates the breakpoints in the components and thus completely solves the boundary problem; otherwise ($\omega \neq 2$), there is always be breakpoints in the components. Specially, when $\omega=1$, although its cosinoidal component is continuous, its sinusoidal component do not include any breakpoints, which is equivalent to partially solving the problem. Therefore, decoded angle waveform is significantly closer to the perfect sawtooth wave compared with original prediction (\cref{fig:original_wave}), but there is still a gap compared with $\omega=2$.

By comparing prediction(dash lines) with ground-truth(solid lines) in \cref{fig:sup_waves}(top), we can find once ground-truth of wrapped value contains breakpoints, its prediction will become significantly worse, and according unwrapped angle, too.
Finally, only $\omega=2$ is the optimal choice that makes $f$ continuous, differentiable, and reversible for rectangular objects.

\subsection{Perspective 3: Experiments}
Although we have analyzed from two different perspectives that $\omega=2$ is the more appropriate angular frequency, the experiment is destined to be the more direct perspective.
We conducted experiments with angular frequencies ($\omega=1,2,4$), as is shown in \cref{tab:freq}. Compared with original KFIoU \cite{yang2023kfiou}, enhanced version with ACM($\omega=1$) get remarkable improvement since sinusoidal component in decomposition of the angle has no breakpoints for rectangles. It is consistent with the phenomenon (the smaller distortion area) observed in the en/decoding waveform diagrams. Moreover, ACM($\omega=2$) eliminates all breakpoints in both two components in decomposition of the angle for rectangles, so it achieves greater improvement. Due to the inability to unwrap the full angular range for rectangles, ACM($\omega=4$) exhibits severe performance degradation, especially for HRSC2016 dataset consisting entirely of large aspect ratio ships. 
When adopted the fusion of two angular frequencies simultaneously ($\omega=2,4$, details in \cref{sec:var} below), compared to $\omega=2$, the results have little  effect  on large aspect ratios objects on HRSC2016 dataset and the results have slightly improved on DOTA dataset. This is because the DOTA dataset contains both large aspect ratio objects and square-like objects.
Overall, AP (especially AP$_{75}$) can benefit a lot from ACM, which verifies our analysis. In following experiments, we adopt mixed angular frequencies ($\omega=2,4$).

\subsection{Extension to Square-like Object}
When the value of the object's width and height are close to each other, the bounding box will become a square-like from rectangle, which possesses stronger symmetry and leads to the period of $B_{angle}$ shrinking from $\pi$ to $\frac{\pi}{2}$. As a result, breakpoints will occur at more locations (i.e., $\frac{\pi}{2}$, $\pi$, and $\frac{3\pi}{2}$). In this case, if we continue to use Angle Correct Module proposed in the previous section, we should set $\omega$ to 4 accordingly, as is shown in Figure~\ref{fig:sup_waves}(bottom). It is worth noting that when $\omega=2$, breakpoints still exist in $f_x$ at $O_{angle}=\frac{\pi}{2}, \pi, \frac{3\pi}{2}$, while $f_y$ suffers from gradient breakpoints at these positions although it is continuous, which is similar to the case of $\omega=1$ for the rectangle. 

\subsection{Generalization for Varied Aspect Ratio}
\label{sec:var}
Considering that the actual scene contains both square-like and rectangular objects, we attempt to use wrapped values with two frequencies (denoted as $f^{(\omega)}$, where $\omega=2, 4$) simultaneously and fuse the unwrapped results to obtain a more accurate angular prediction. Similar strategies can also be found in previous work \cite{zhu2020adaptive,yu2023psc}. For boxes rotated within $[0, \frac{\pi}{2})$, both $f^{(2)}$ and $f^{(4)}$ can unwrap correct angles. For boxes rotated within $[\frac{\pi}{2}, \pi)$, $f^{(2)}$ still unwraps correctly, while $f^{(4)}$ 's unwrapped angles will be offset by one decoding-period $\frac{\pi}{2}$ to fall in $[0, \frac{\pi}{2})$. Therefore, ideally the difference between $\theta^{(2)} \in [0, \pi)$ and $\theta^{(4)} \in [0, \frac{\pi}{2})$ could only be $0$ or $\frac{\pi}{2}$, but it only affects rectangle ($T=\pi$) and not square-like ($T=\frac{\pi}{2}$) in both training and inference phase. Note that $f^{(2)}$ suffers from breakpoints only for square-like rather than rectangle, and $f^{(4)}$ is immune to breakpoints for both rectangle and square-like, which just fails to correctly determine period range belonging to angles. Thus we can utilize coarse $\theta^{(2)}$ to correct the period range of fine $\theta^{(4)}$ as follows, where relaxation condition outside the parentheses are adopted in practice due to the independent errors of $f^{(2)}$ \& $f^{(4)}$. Finally, we use this fusion strategy to adapt objects with varied aspect ratio.
\begin{equation}
\begin{aligned}
\theta = 
\begin{cases}
\theta^{(4)} + \frac{\pi}{2} &, if \ \theta^{(2)} - \theta^{(4)} > \frac{\pi}{4} \ (\theta^{(2)} - \theta^{(4)} = \frac{\pi}{2})\\
\theta^{(4)} &, if \ \theta^{(2)} - \theta^{(4)} \leq \frac{\pi}{4} \ (\theta^{(2)} = \theta^{(4)})
\end{cases} 
\end{aligned}
\end{equation}
where the inequality condition (e.g. $\theta^{(2)} - \theta^{(4)} > \frac{\pi}{4}$) is just a relaxed version of the equality condition (e.g. $\theta^{(2)} - \theta^{(4)} = \frac{\pi}{2}$). The latter is just the judgment condition in the ideal state, while the former is actually adopted in practice, which brings better numerical stability.

\end{document}